\documentclass[letterpaper, 10 pt, conference]{ieeeconf}  % Comment this line out if you need a4paper
\usepackage{amsmath}
\usepackage{amssymb}
\usepackage{graphicx}
\usepackage{multirow}
\usepackage{colortbl}
\usepackage{subcaption}
\usepackage{tabularx}
\usepackage{bm}
\newcommand{\lang}{\mathcal{L}}
\newcommand{\policypi}{\pi^P}
\newcommand{\langModel}{\pi^L}
\newcommand{\codeModel}{\pi^C}
\newcommand{\MethodName}{\textbf{\emph{T2D}}}
\newcommand{\MethodNames}{\textbf{\emph{T2D }}}
\newcommand{\llst}{\mathcal{M}}
\newcommand{\translator}{\llst}
\newcommand{\primary}{\mathcal{R}_{\text{P}}}
\newcommand{\auxiliary}{\mathcal{R}_{\text{aux}}}
\newcommand{\reward}{\mathcal{R}}
\newcommand{\lowstate}{\mathcal{S}}
\newcommand{\lowaction}{\mathcal{A}}

\newcommand{\abstractstates}{Q}
\newcommand{\abstractstate}{q}
\newcommand{\stateshistory}{\mathcal{H}^{Q}}
\newcommand{\statehistory}{\mathcal{H}^{q}}
\newcommand{\transition}{\mathcal{T}}
\newcommand{\behaviortransition}{\delta}
\newcommand{\codepi}{\pi^{C}}
\newcommand{\timestep}{T}
\newcommand{\event}{E}
\newcommand{\guard}{G}
\newcommand{\update}{U}
\newcommand{\alphabet}{\Sigma}

\newcommand{\initial}{\abstractstate_0}

\newcommand{\fsmconclusion}{F}

\IEEEoverridecommandlockouts                              % This command is only needed if 
\title{\LARGE \bf
\emph{Text-to-Drive}: Diverse Driving Behavior Synthesis \\ via Large Language Models
}

\author{Phat Nguyen$^{1}$, Tsun-Hsuan Wang$^{2}$, Zhang-Wei Hong$^{2}$, Sertac Karaman$^{3}$, Daniela Rus$^{2}$% <-this % stops a space
\thanks{$^{1}$UMass Amherst, $^{2}$MIT CSAIL, $^{3}$MIT LIDS.}
}

\usepackage[table,xcdraw]{xcolor}
\usepackage{tikz}

\usepackage[noframe]{showframe}
\usepackage{framed}

\usepackage{fancyvrb}

\renewenvironment{shaded}{%
  \MakeFramed{\advance\hsize-\width \FrameRestore\FrameRestore}}%
 {\endMakeFramed}
\definecolor{shadecolor}{gray}{0.8}

\usepackage{hyperref}
\renewcommand{\sectionautorefname}{Sec.}
\renewcommand{\subsectionautorefname}{Sec.}
\renewcommand{\subsubsectionautorefname}{Sec.}
\definecolor{mylinkcolor}{RGB}{31, 96, 153}
\hypersetup{
    colorlinks=true,    % no colored links
    pdfborder={0 0 0},   % a box around links (set to 0 0 0 to remove the border)
    allcolors=mylinkcolor,
}
\usepackage{cleveref}

\begin{document}

\maketitle
\thispagestyle{empty}
\pagestyle{empty}
% \begin{figure*}[t]
%         \centering
%         \includegraphics[width=\textwidth]{Figures/merge_trajectories.png}
%         \caption{Trajectory Visualization on the Merge map.}
%         \label{fig:header_photo}
% \end{figure*}

%%%%%%%%%%%%%%%%%%%%%%%%%%%%%%%%%%%%%%%%%%%%%%%%%%%%%%%%%%%%%%%%%%%%%%%%%%%%%%%%
\begin{abstract} 
        Generating varied scenarios through simulation is crucial for training and evaluating safety-critical systems, such as autonomous vehicles. Yet, the task of modeling the trajectories of other vehicles to simulate diverse and meaningful close interactions remains prohibitively costly. Adopting language descriptions to generate driving behaviors emerges as a promising strategy, offering a scalable and intuitive method for human operators to simulate a wide range of driving interactions. However, the scarcity of large-scale annotated language-trajectory data makes this approach challenging. To address this gap, we propose Text-to-Drive (T2D) to synthesize diverse driving behaviors via Large Language Models (LLMs). We introduce a knowledge-driven approach that operates in two stages. In the first stage, we employ the embedded knowledge of LLMs to generate diverse language descriptions of driving behaviors for a scene. Then, we leverage LLM's reasoning capabilities to synthesize these behaviors in simulation. At its core, T2D employs an LLM to construct a state chart that maps low-level states to high-level abstractions. This strategy aids in downstream tasks such as summarizing low-level observations, assessing policy alignment with behavior description, and shaping the auxiliary reward, all without needing human supervision. With our knowledge-driven approach, we demonstrate that T2D generates more diverse trajectories compared to other baselines and offers a natural language interface that allows for interactive incorporation of human preference. Please check our website for more examples: \href{https://text-to-drive.github.io/}{here}
\end{abstract}

%%%%%%%%%%%%%%%%%%%%%%%%%%%%%%%%%%%%%%%%%%%%%%%%%%%%%%%%%%%%%%%%%%%%%%%%%%%%%%%%
\section{INTRODUCTION}

Simulators have emerged as an effective tool for training and evaluating safety-critical systems, such as autonomous vehicles. They provide opportunities to synthesize novel data for training, expose methods to edge cases that are otherwise not available in public driving datasets, and offer a cost-effective method of simulating close interactions that are otherwise costly or impractical to capture in real-world settings. Their utility extends further to in-simulation validations and enables detailed studies that are difficult to observe directly. 

Despite their advantages, current simulators face significant challenges in controlling the behaviors of surrounding vehicles and scaling these interactions. Adding varied driving behaviors to the simulation can facilitate comprehensive testing across diverse driving behavior profiles. This addresses the inherent bias found in driving datasets, used by data-driven simulators, which tend to be limited in scope and curated from a narrow selection of geographic areas.

One promising direction to overcome these limitations is by extending the capabilities of foundational models into simulators. A knowledge-driven approach that utilizes the embedded knowledge of Large Language Models (LLMs) to curate comprehensive and diverse driving scenarios, eliminates the need for exhaustive manual scripting of potential interactions. Furthermore, utilizing natural language to control the generation of these scenarios presents an intuitive method to specify desired behavior trajectories. This technique allows for generating driving scenarios based on language descriptions, making it easier for human operators to curate meaningful test cases. A notable application of this strategy involves connecting to textual data, such as accident reports, to ground simulations in real-world contexts. Our work adopts this knowledge-driven approach, drawing on rich knowledge sources to generate diverse driving scenarios. This method can be complementary to data-driven simulators, which rely only on human-driving data.

\begin{figure}[t]
        % Remove the \centering command
        \centering
        \includegraphics[width=\linewidth]{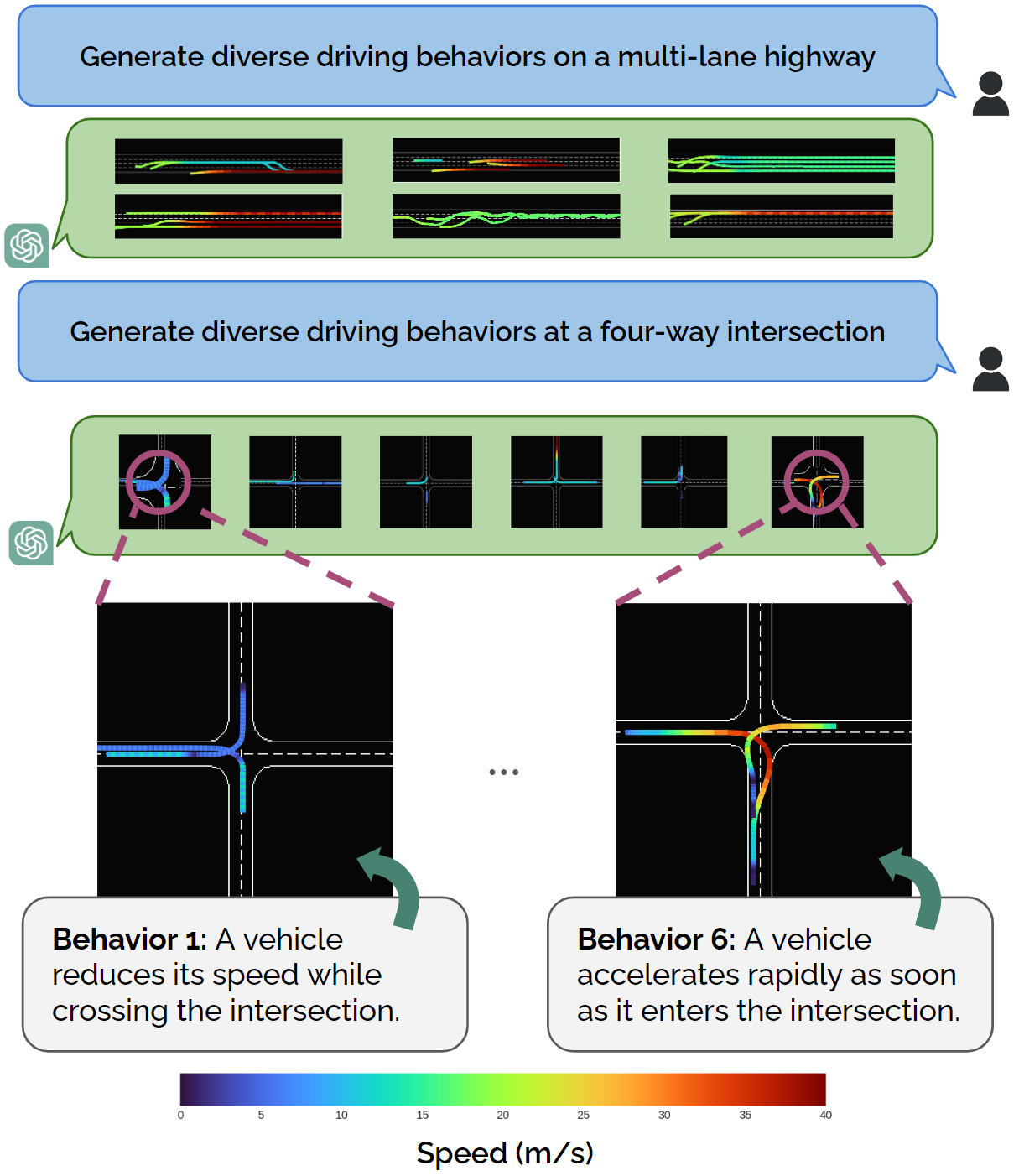}
        \caption{Given a scene description, \textbf{\emph{T2D}} leverages Large Language Models to generate diverse descriptions of driving behaviors and then synthesizes them in simulation.}
        \vspace{-0.5cm}
        \label{fig:teaser}
\end{figure}
\begin{figure*}[t]
        \centering
        \includegraphics[width=0.97\linewidth]{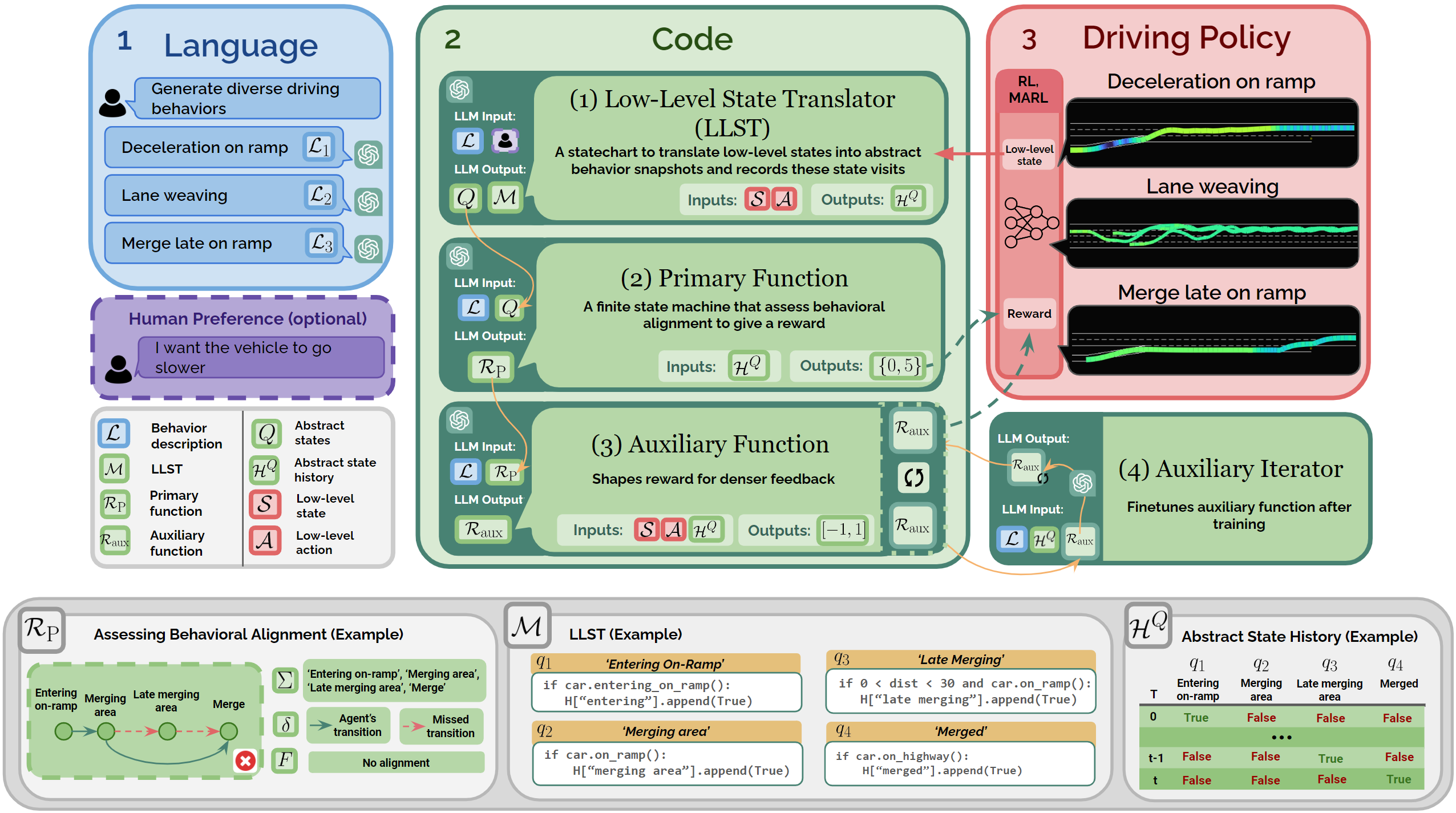}
        \caption{\textbf{Overview.} 
        \emph{Left}: First, an LLM generates diverse descriptions of driving behaviors, which can incorporate human preferences through a natural language interface.
        \emph{Middle}: Next, an LLM generates a low-level state translator (LLST), primary function, and auxiliary function from a description of a driving behavior. The LLST translates low-level states to abstract states (see example in \emph{bottom middle} block) and then records their state visit history (see example in \emph{bottom right} block). The primary function gives a reward only when the vehicle exhibits the target behavior, using a finite-state machine for formal verification of behavior emergence (see example in \emph{bottom left} block). The auxiliary function provides rewards for reaching intermediate states and can be iteratively updated.
        \emph{Right}: Finally, we employ a standard multi-agent RL framework to train a driving policy using the primary and auxiliary functions as guidance.}
        \vspace{-0.5cm}
        \label{fig:overview}
\end{figure*}

In this paper, our research aims to answer the question: \emph{How can we translate a diverse set of language descriptions of driving behaviors into a corresponding set of behaviorally diverse policies for simulation?} To address this, we introduce \MethodName, a knowledge-driven method for simulation that utilizes LLMs to generate diverse language descriptions of driving behaviors and then synthesizes them in simulation. Given a behavior description, \MethodNames generates a mapping of low-level states (e.g: vehicle position, heading, speed) to high-level abstractions (e.g. ``on the on-ramp'', ``near the end of on-ramp'', and ``merged''). By leveraging this abstract state representation, transitions are defined to capture the temporal dynamics of the behavior, effectively embodying temporal logic. Such a framework not only enhances the capacity for reasoning about behavioral alignment but also creates abstract summaries of low-level observations. The ability to assess the behavioral alignment is used as a primary reward function to guide the driving policy, while our auxiliary function is used to improve exploration efficiency. The abstract summaries inform the LLM whether and how to adjust the auxiliary function after training. This iterative process introduces new incentives for exploring new states and penalties for unwanted behaviors, thereby aligning the policy more closely with the desired behavior. We demonstrate that \MethodNames maintains the behavioral context across natural language, code, and driving policy, enabling accurate simulation of the driving behavior. Additionally, \MethodNames surpasses baselines in generating diverse trajectories and offers a natural language interface to embed human preferences into the driving trajectories. Using \MethodName, we generated 18 driving behaviors from language descriptions. To this end, we make the following key contributions:

\begin{itemize}
        \item {We introduce \MethodName, a knowledge-driven method for simulation that enables (i) text-to-driving behavior synthesis, and (ii) diverse driving behavior generation.}
        \item {Our method facilitates the use of LLM-based reasoning by encapsulating the logic in state machines. This facilitates complex policy training processes such as: (i) summarizing low-level observations, (ii) reasoning about behavioral alignment, and (iii) iteratively updating the auxiliary function, without any human supervision.}
        \item {We demonstrate our method effectively retains the behavioral context across natural language, code, and driving policy, enabling it to simulate a driving behavior from a description. Additionally, \MethodNames not only generates more diverse trajectories compared to baselines but also offers a language interface to integrate human preferences into driving simulations.}
\end{itemize}

% Diversity while maintaining realism

\section{RELATED WORK}

\subsection{Driving Simulators}

% Model-based simulators
        % CARLA
% NOTE: NOT DONE !!!!!!!!!!!!!!!!!!!
Collecting data for specific scenarios proves challenging, emphasizing the need to test safety-critical systems in controlled environments. This necessity has spurred the development of model-based simulators \cite{Dosovitskiy17, nvidia2022drivesim}, capable of modeling real-world physics and constructing photorealistic environments. Continual improvements have integrated tools like Scenic \cite{Fremont_2019}, enabling the creation of complex traffic scenarios using compact English-like syntax. However, these simulators often struggle with a sim-to-real domain gap \cite{tobin2017domain}. In response, data-driven simulators \cite{amini2022vista, amini2020learning, manivasagam2020lidarsim, gulino2023waymax, li2022metadrive} can bridge this gap by leveraging real-world driving data \cite{caesar2022nuplan, sun2020scalability, liao2022kitti360} to reconstruct real-world scenes and synthesize novel views. These simulators, however, lack the generative control that model-based simulators offer. Our work addresses this issue by extending the generative capabilities of foundation models into simulators, specifically those that are equipped with a reinforcement learning (RL) interface \cite{gulino2023waymax,highway-env}.

\subsection{Behavior Generation}

% IDM
% Unsupervised Skill Acquisition
        % DIAYN: https://arxiv.org/abs/1802.06070
        % DADS
        % 3: Trajectory diversity: https://proceedings.mlr.press/v139/lupu21a/lupu21a.pdf
        % Behaviorally Diverse Traffic Simulation via Reinforcement Learning: https://arxiv.org/abs/2011.05741
\textbf{Diverse Skills.} Unsupervised skill acquisition algorithms such as DIAYN \cite{eysenbach2018diversity} and DADS \cite{sharma2020dynamicsaware} have demonstrated their ability to learn diverse skills in unsupervised settings. These have further inspired methods aimed at enhancing trajectory diversity in RL \cite{pmlr-v139-lupu21a} and simulating diverse traffic behaviors \cite{shiroshita2020behaviorally}. However, despite their ability to generate diverse actions, their methods cannot be directed through textual guidance or incorporate human preference.

% Generating without language + NN to learn from real-world trajectory datasets
        % L. Bergamini, Y. Ye, O. Scheel, L. Chen, C. Hu, L. Del Pero, B. Osinski, H. Grimmett, and P. On- ´ druska, “Simnet: Learning reactive self-driving simulations from real-world observations,” in 2021 IEEE International Conference on Robotics and Automation (ICRA), pp. 5119–5125, IEEE, 2021.
        %  S. Suo, S. Regalado, S. Casas, and R. Urtasun, “Trafficsim: Learning to simulate realistic multi-agent behaviors,” in Proceedings of the IEEE/CVF Conference on Computer Vision and Pattern Recognition, pp. 10400–10409, 2021.
        % D. Rempe, J. Philion, L. J. Guibas, S. Fidler, and O. Litany, “Generating useful accident-prone driving scenarios via a learned traffic prior,” in IEEE Conf. on Computer Vision and Pattern Recognition, 2022.
        % TrafficGen: https://metadriverse.github.io/trafficgen/
        % SceneGen: https://arxiv.org/abs/2101.06541
        % BITS: Bi-level Imitation for Traffic Simulation: https://arxiv.org/abs/2208.12403
% Using diffusion: https://arxiv.org/pdf/2401.00391.pdf
% Retrieval:
        % RealGen: https://arxiv.org/abs/2312.13303 
% NOTE: NOT DONE !!!!!!!!!!!!!!!!!!!
        % data-driven
\textbf{Controllable Traffic Generation.} Recent advancements in data-driven traffic generation methods have employed neural networks to synthesize new scenarios \cite{bergamini2021simnet,suo2021trafficsim,rempe2022generating,feng2023trafficgen, tan2021scenegen, xu2022bits}, offering avenues for more realistic traffic modeling. Complementing these efforts, research has expanded into multiple directions: learning latent representations from driving datasets \cite{ding2023realgen, li2023scenarionet}, incorporating human preferences with reinforcement learning \cite{cao2023reinforcement}, and exploring generative scenarios through diffusion models \cite{chang2023controllable, cao2023reinforcement}. These efforts collectively enhance the controllability of simulated environments, yet they cannot take text descriptions of driving behaviors as inputs. The necessity of manual labeling for driving behaviors in recent research \cite{ding2023realgen, sima2023drivelm} further highlights this limitation.

% Language Condition Generation
        % LCTGen: https://arxiv.org/abs/2307.07947
        % Language-Guided Diffusion: zhong2023languageguided

More recently, language-conditioned traffic generation has been explored in \cite{tan2023language,zhong2023languageguided, roh2019conditional}. \cite{tan2023language} leverages LLMs to translate textual descriptions of traffic scenes directly into driving trajectories. However, while their approaches focus on low-level trajectory generation, ours explores the generation of diverse high-level behaviors such as ``tailgating''. In this work, we build upon the capabilities of LLMs for zero-shot generation of reward functions \cite{ma2023eureka, song2023selfrefined, yu2023language}. Our research explores this capability further and extends it to diverse driving behaviors for simulation, especially in scenarios lacking a ground-truth fitness function. 

% \section{PRELIMINARIES}

% A driving behavior policy, $\policypi$, determines the agent's action choices by generating a probability distribution, $\lowaction$, over a discrete action space $\lowaction \in \mathbb{R}^{5}$ such that $\lowaction \sim \policypi(.|\lowstate)$, where $\lowstate \in \mathbb{R}^{25}$ represents the agent's current low-level state. The action space consists of five discrete actions: accelerate, decelerate, idle, left lane change, and right lane change. 

% During a rollout, a behavior policy creates a scenario characterized by the driving trajectory $\tau \in \mathbb{R}^{T \times N \times 6}$ and action distribution trajectory $\tau_{action} \in \mathbb{R}^{\timestep \times N \times 5}$, where $\timestep$ is the number of timesteps, $N$ is the number of controlled vehicles. Each agent's state trajectory is composed of the parameters $[p_x, p_y, v_x, v_y, h_x, h_y]$ with position $p_x$, $p_y$, velocity $v_x$, $v_y$, and heading $h_x$, $h_y$. The action distribution trajectory represents the sequence of action choices guided by $\policypi$.

\section{METHOD}
The goal of our method is to generate diverse driving behaviors from textual inputs through a knowledge-driven approach, structured into two main stages. In the first stage, we utilize the vast knowledge embedded in LLMs to generate language descriptions of diverse driving behaviors (\autoref{sec:language_gen}). In the second stage, as we transition from these descriptions to driving policies, three key steps are undertaken. First, we used an LLM to generate a low-level state translator (\autoref{sec:llst}). We detail how this is used to translate low-level trajectories to an abstract code representation. Then, to evaluate the alignment of driving policies with desired behaviors, we introduce the primary reward function (\autoref{sec:fitness}), which, together with our auxiliary function, provides reward guidance to the driving policy (\autoref{sec:auxiliary}). Subsequently, we detail an iterative approach that employs LLMs to adjust the auxiliary function after each training. Finally, we employ a multi-agent RL framework to train a driving policy (\autoref{sec:policy_learning_method}). An overview of our method is shown in \autoref{fig:overview}.

\subsection{Generating Behavior Descriptions}
\label{sec:language_gen}
In our knowledge-driven approach, we use \texttt{gpt-4}'s zero-shot generation capability to generate descriptions of diverse driving behaviors, $\lang \sim \langModel(.|\text{scene})$ from a concise description of a scene. For each scene type -- intersections, merges, highways -- we generated 50 descriptions of driving behaviors. We then selected a smaller, representative subset of six behaviors per scene for simulation to provide a detailed analysis of all generated behaviors.

\subsection{Retrieval from Environment}
\label{sec:rag}
We enhance our code generation model, $\codeModel$, with retrieval augmented generation (RAG) to provide sufficient context about the simulation environment. This involves segmenting the source code using an abstract syntax tree (AST), embedding the code using a text embedding model (\texttt{text-embedding-ada-002}), and storing the embeddings in a database. To preserve the dependencies between the code segments, we use ctags to generate a repository map. We used LangChain to implement our RAG framework.

During the retrieval process, we use the behavior description $\lang$ to query the embedding database and retrieve relevant code segments. We use this code and a repository map as context for $\codeModel$. By making the source code accessible, we enrich the code generation model with APIs about the low-level state. For instance, in addressing the behavior ``accelerative merging on-ramp'', the model can utilize the attribute \texttt{car.speed} and access functions \texttt{car.on\_ramp()}. This retrieval ensures that the low-level observations of the driving policy are available in the code space.

\begin{figure*}[t]
        \centering
        \begin{subfigure}[b]{.32\linewidth}
            \centering
            \includegraphics[width=\linewidth]{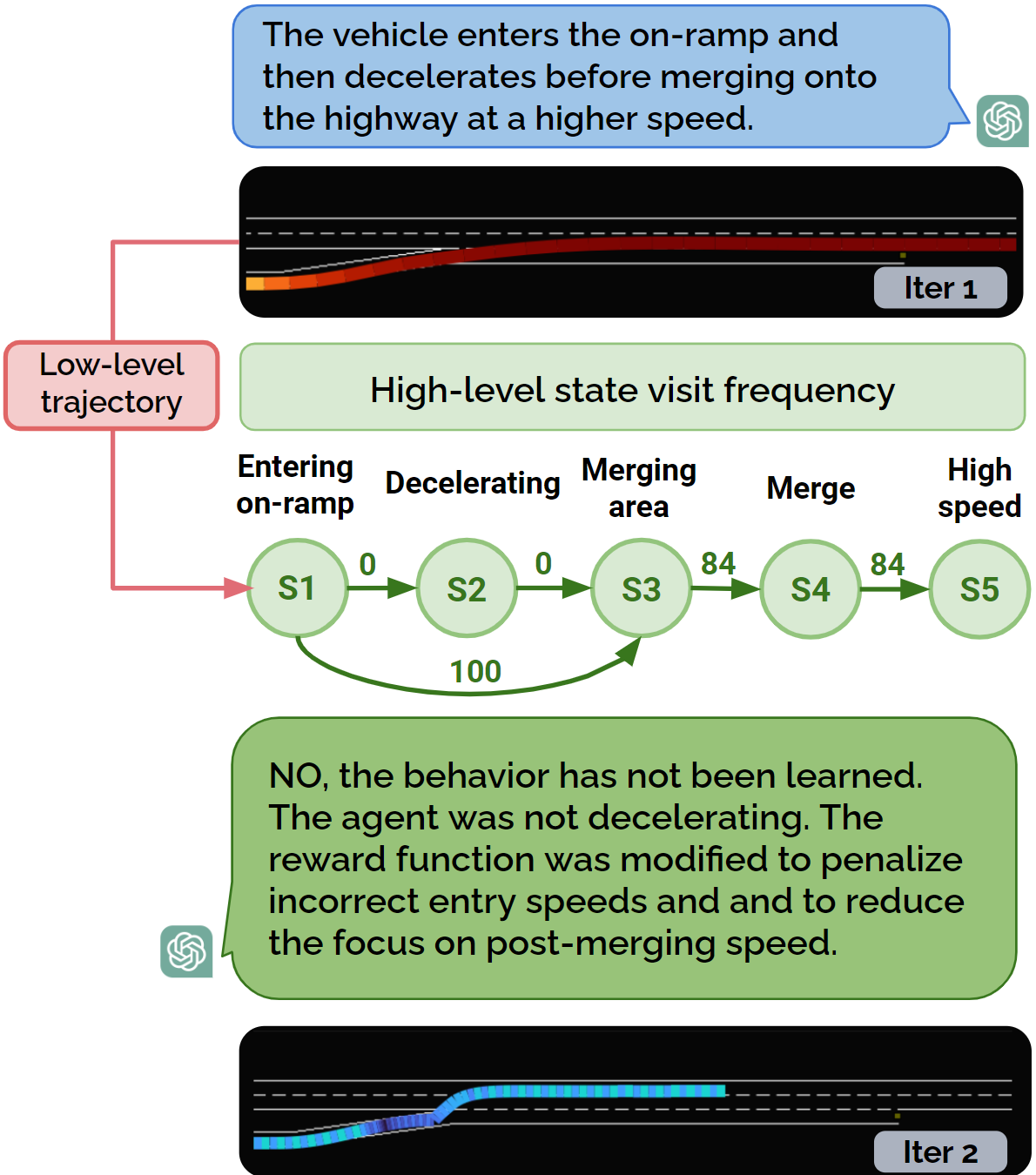}
        \end{subfigure}\hfill
        \begin{subfigure}[b]{.66\linewidth}
            \centering
            \includegraphics[width=\linewidth]{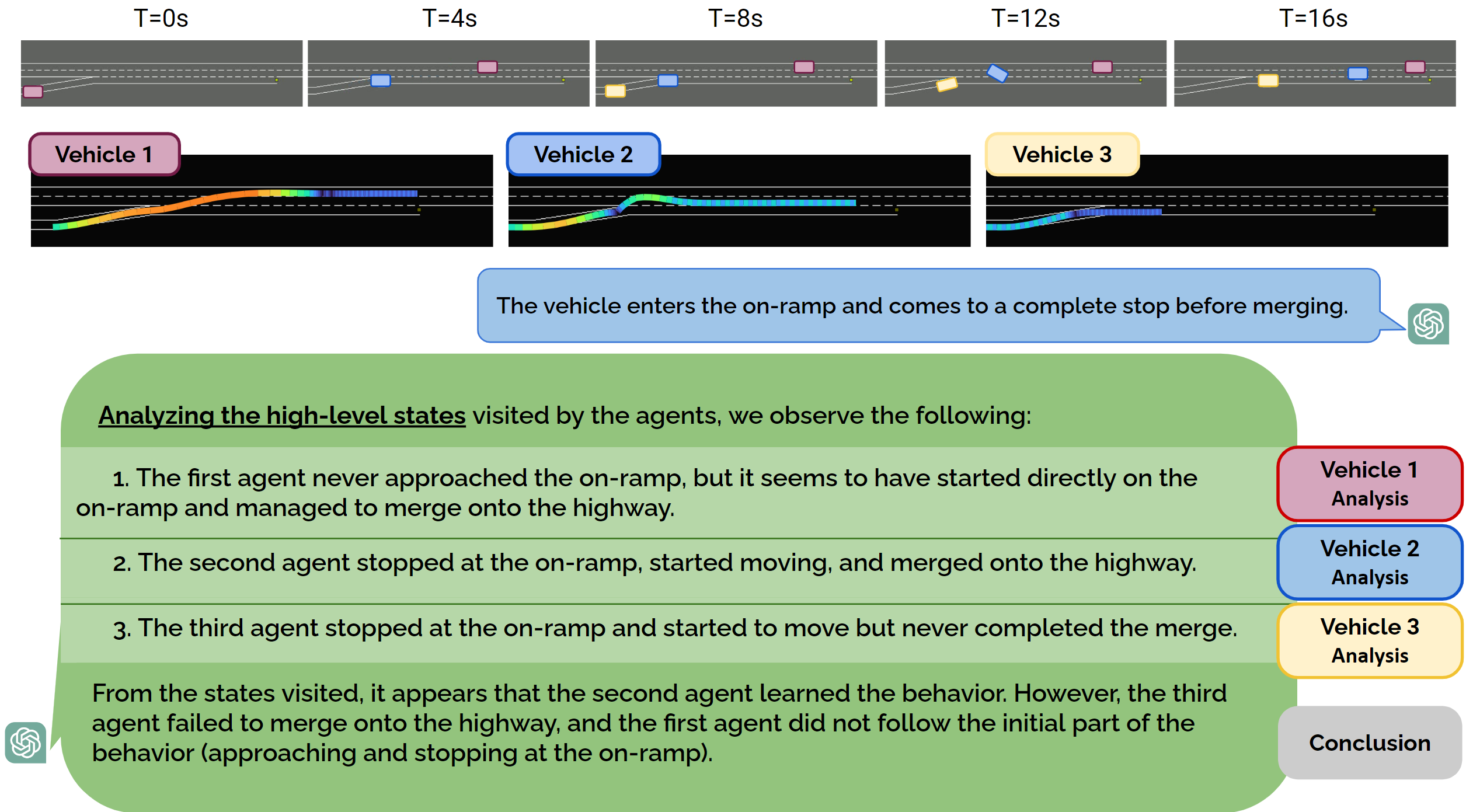}
        \end{subfigure}
        \caption{\emph{Left}: The auxiliary iterator LLM analyzes the policy after training to decide whether and how to adjust the auxiliary function based on the history of abstract state visits. \emph{Right}: The right figure illustrates the LLM's reasoning process, where it reads a high-level behavior sequence, analyzes it, and then provides an accurate summary of the low-level trajectories.}
        \vspace{-0.5cm}
        \label{fig:iterator_state_summary}
\end{figure*}

\subsection{Low-Level State Translator (LLST)}
\label{sec:llst} 

\newcommand{\state}{\mathcal{S}}

Given a behavior description $\lang$, our code generation model $\codeModel$, generates a low-level state translator, $\translator \sim \codeModel(.|\lang)$. The state translator $\translator$, has three primary responsibilities: it (1) decomposes the behavior, (2) maps lower-level states to abstract states, and (3) records abstract state visits. First, $\translator$ decomposes the behavior into abstract states, $\abstractstates$. Each abstract state, $\abstractstate \in \abstractstates$, captures an essential aspect of the driving behavior. For example, in the case of ``merging late on the on-ramp'', $\abstractstate$ could be any of ``on the on-ramp,'' ``merging,'' and ``near the end of the on-ramp'' as illustrated in \autoref{fig:overview}. This decomposition yields four advantages: firstly, it discretizes the behavior into critical phases to capture different aspects of the behavior; secondly, it constrains $\abstractstates$ to be relevant to the target behavior; thirdly, by merging the initial advantages, it imparts clear, objective guidance which makes the generation of $\translator$ more consistent and reliable; and lastly, it allows the state name to be in the language domain. 

        % MAPPING LOW TO HIGH STATES:
Moreover, $\translator$ is constructed by the LLM as a state chart. We define $\translator$ as a tuple $\translator = (\abstractstates, \transition, E, U, G)$, where $\transition$ is the set of transitions triggered by an event $\event$, conditioned on a guard in $\guard$, and results in an update action from $\update$. The events in $\event$ represent low-level changes within the driving environment, such as speed, position, and heading. The guards in $\guard$ are boolean functions that return true under certain conditions in the low-level state. We exploit the code-generating capabilities of LLMs to achieve semantic alignment between guard conditions and abstract state names. This step leverages the code-generation model, $\codeModel$, with our RAG framework. Specifically, we used the \texttt{gpt-4-1106-preview} variant \cite{OpenAI2023GPT4} and set the temperature to $0.2$. This decision aims for a more deterministic output due to the objective nature of the task.

The translator's update action from $\update$, results in an update in the state history dictionary $\stateshistory$, that keeps a historical record of all abstract state visits. With each timestep, it appends a boolean value indicating whether a state has been visited. The state history dictionary $\stateshistory$ over a rollout of $\timestep$ timesteps can be represented as $\stateshistory : \abstractstates \rightarrow \{\text{true}, \text{false}\}^{T}$. Then $\stateshistory(\abstractstate) = \statehistory$, is the visit history associated with $\abstractstate$. \emph{This enables both the ability to identify current state occupancy and track state visits and transitions.} These characteristics are extremely effective at summarizing low-level observations of the driving policy back to the code and language space (see \autoref{fig:iterator_state_summary}) which we utilized in the primary function (\autoref{sec:fitness}) and iterator (\autoref{sec:iterator}).

\subsection{Primary Reward Function}
\label{sec:fitness} 
In this section, we introduce the primary function $\primary$, generated by an LLM, which takes $\stateshistory$ as input and returns a reward, $\primary : \stateshistory \rightarrow \{0, 5\}$. This function serves two purposes: first, it assesses the behavioral alignment of the driving policy $\policypi$ with the target behavior $\lang$; second, it awards a large reward when the vehicle demonstrates the target behavior. To generate $\primary \sim \codeModel(.|(\abstractstates, \lang))$, we combine the abstract state names $\abstractstates$, and behavior description $\lang$, as inputs to $\codeModel$. The LLM constructs $\primary$ as a finite-state machine (FSM) that models the target behavior $\lang$. This FSM can be described as a tuple, $\primary = (\abstractstates, \alphabet, \behaviortransition, \initial, \fsmconclusion)$; where $\alphabet = \{\abstractstate | \abstractstate \in \abstractstates \}$ is the input alphabet, $\behaviortransition: \abstractstates \times \alphabet \rightarrow \abstractstates$ is the transition function, $\initial$ is the initial state, and $\fsmconclusion$ is the accepting states, indicating the target behavior is achieved.

The abstract state of $\primary$ utilizes $\abstractstates$ from the state translator, $\translator$. We regard each $\abstractstate \in \abstractstates$ from $\translator$ as a code abstraction representing a behavior snapshot. For instance, the state ``end of on-ramp'' is an abstraction for the conditions \texttt{$0 < $ car.headway() $<30$} and \texttt{car.on\_ramp()}. The behavioral transition function $\delta$, then adds transitions between abstract states to capture the temporal dynamics of the target behavior. This is particularly useful for driving behaviors such as ``late merging'',  which requires visiting ``end of on-ramp'' before transitioning to the ``merge'' state. The formal structure of the FSM, generated by the LLM, provides a framework for verifying the abstract behavior sequences given by $\stateshistory$. This strategy encapsulates the LLM's reasoning into a compact FSM that can be accessed after LLM inference. This structured format \emph{enables an offline application that utilizes LLM's reasoning to assess behavioral alignment during the training of the driving policy}. For a visual illustration of the FSM, refer to \autoref{fig:overview}. In our implementation, we set the environment minimum speed to 0 so that the vehicle could not reverse to avoid complex generation of $\primary$.

We then use the FSM to give a reward of $5$ to the vehicle upon reaching the accepted states, $\fsmconclusion$. This reason will be evident in the next section (\autoref{sec:auxiliary}). An important nuance of our method is its bi-functional relationship between the state translator and the primary function. The state translator abstracts low-level observations of the driving policy into the code domain, while $\primary$ evaluates these abstractions in the code domain and gives a reward to guide the driving policy $\policypi$. This reciprocal relationship encourages behavioral consistency across different spaces.

\subsection{Auxiliary Reward Function}
\label{sec:auxiliary} 
Although the primary function $\primary$, is used to guide the driving policy $\policypi$, the rewards from $\primary$ are too sparse. To improve exploration efficiency, we use an auxiliary function, $\auxiliary : \lowstate, \lowaction, \stateshistory \rightarrow [-1,1]$, that takes low-level state $\lowstate$, action $\lowaction$, and abstract state history $\stateshistory$, as inputs and returns a reward. The purpose of providing more input to the auxiliary function is to enable reward guidance for both low-level and abstract states. Our auxiliary function is generated using the code generation model $\codeModel$, with $\primary$ and $\lang$ as inputs, $\auxiliary \sim \codeModel(.|(\primary, \lang))$.  We use the same RAG framework and LLM model with a temperature of $0.7$ to allow for more creative reward shaping. 

The auxiliary function purpose is to provide denser rewards for reaching intermediate states, while simultaneously injecting the behavioral context into the driving policy. This is particularly useful for complex behaviors characterized by numerous transitions where the primary reward function offers limited guidance. For instance, the behavior ``merging from a complete stop'' requires the vehicle to navigate a series of actions: enter the on-ramp, stop, move again, and then merge. Using the auxiliary reward, we successfully trained 18 driving behaviors (see \Cref{fig:intersection_traj_visualization,fig:merge_traj_visualization,fig:highway_traj_visualization}).

The rewards from our $\auxiliary$ are normalized to the range $[-1,1]$. Empirically, we find that as the driving policy learns the behavior, the auxiliary function's influence naturally diminishes as the primary function's larger rewards become more dominant. This makes our method more robust to different generations of the auxiliary function.

\subsection{Auxiliary Function Iterator}
\label{sec:iterator}
\newcommand{\codeiter}{\pi_{\mathcal{I}}^C}

A common challenge of reward shaping is that it can generate unintended behaviors. To address this, we generate a new iteration of the auxiliary function using a code generation model, $\auxiliary' \sim \codepi(.|(\auxiliary, \statehistory, \lang))$; where $\auxiliary'$ is the new auxiliary function, and $(\auxiliary, \statehistory, \lang)$ are the inputs to the LLM. The abstract state history $\statehistory$, as an input to $\codepi$, provides an informative abstract summary of the low-level trajectories. This summary provides the LLM with high-level insights into a rollout, allowing it to adjust the reward incentive structure accordingly. This iterative process introduces new incentives for exploring new states and penalties for unwanted behaviors, thereby aligning the policy more closely with the desired behavior. By incorporating high-level observations into the iterative process, we proactively mitigate risks associated with unsafe reward-shaping practices \cite{knox2022reward}. For a visual illustration of this process, see the left figure in \autoref{fig:iterator_state_summary}.

\subsection{Learning a Driving Policy.} 
\label{sec:policy_learning_method} 
We employed a multi-agent implementation of the Advantage Actor-Critic algorithm (MAA2C) \cite{mnih2016asynchronous} to learn the driving policies $\policypi$. In this setup, each agent independently learns using a concurrent training strategy in a cooperative environment under partial observation conditions.

\begin{figure}
        \centering
        
        % Top subfigure with an image
        \begin{subfigure}[b]{\linewidth}
                \centering
                \includegraphics[width=\linewidth]{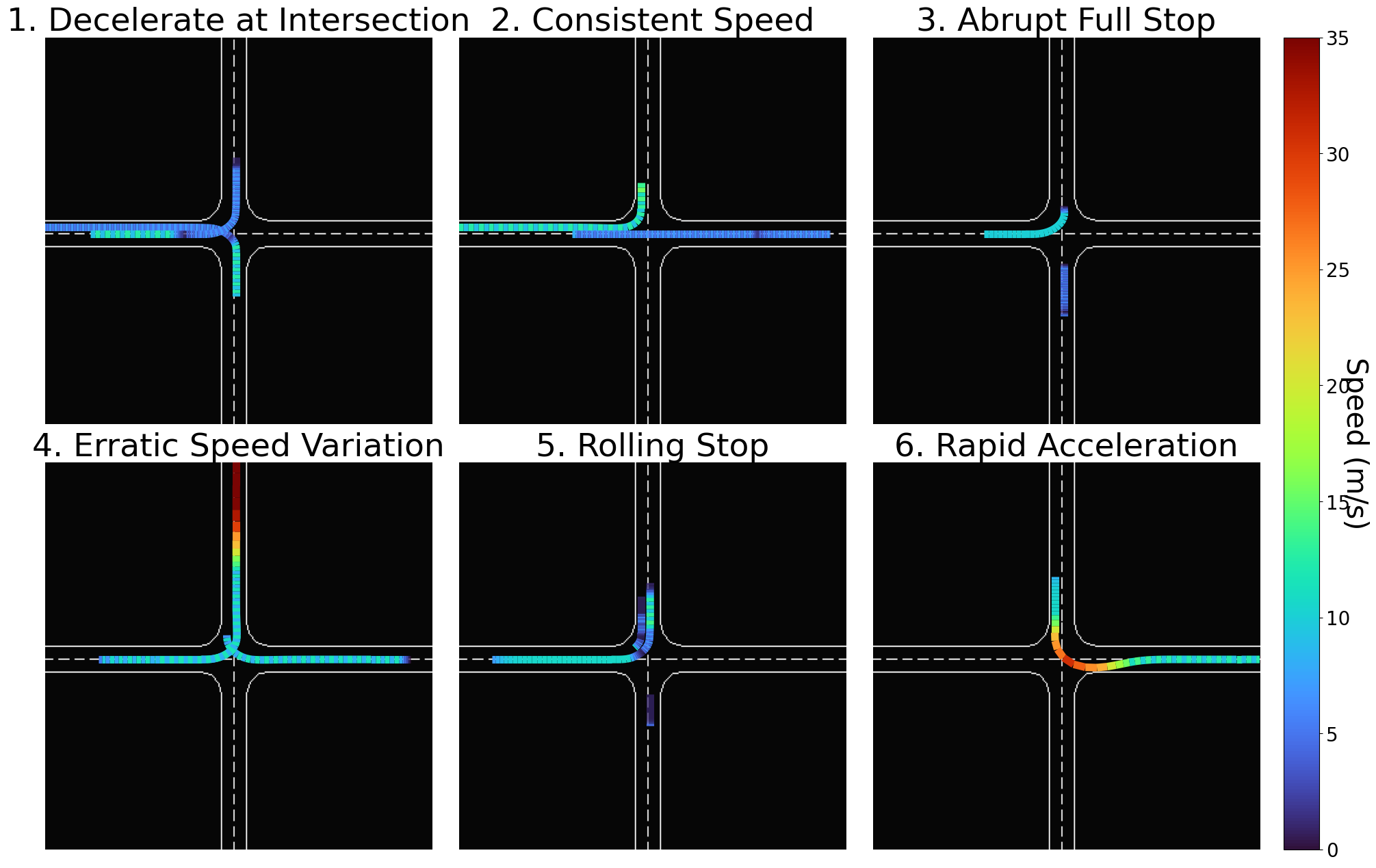}
                % \caption{Speed profile trajectory of driving behaviors at an intersection.}
                \label{fig:intersection_traj_visualization}
        \end{subfigure}
        
        % Add some space between the two subfigures
        \vspace{-0.2cm}
        
        % Bottom subfigure with a table
        \begin{subfigure}[b]{\linewidth}
          \centering
          \resizebox{\linewidth}{!}
          {\begin{tabular}{lccc}
                \hline
                \multicolumn{1}{c}{Behaviors}      & \begin{tabular}[c]{@{}c@{}}Emergence Rate\\ (\%)\end{tabular} & \begin{tabular}[c]{@{}c@{}}Collision Rate\\ (\%)\end{tabular} & \begin{tabular}[c]{@{}c@{}}Avg. Speed\\ (m/s)\end{tabular} \\ \hline
                \rowcolor[HTML]{EFEFEF} 
                1. Decelerate through intersection    & 56.67                                                         & 40.00                                                         & 7.42                                                       \\
                2. Consistent speed crossing          & 63.33                                                         & 20.00                                                         & 5.73                                                       \\
                \rowcolor[HTML]{EFEFEF} 
                3. Abrupt full stop at intersection   & 56.67                                                         & 13.33                                                         & 1.06                                                       \\
                4. Erratic speed                      & 93.33                                                         & 26.67                                                         & 5.60                                                       \\
                \rowcolor[HTML]{EFEFEF} 
                5. Rolling stop at intersection       & 73.33                                                         & 30.00                                                         & 5.10                                                       \\
                6. Rapid acceleration at intersection & 76.67                                                         & 43.33                                                         & 21.24                                                      \\ \hline
        \end{tabular}}
        % \caption{Human-annotated emergence rate and driving characteristics}
        \label{tab:intersection_characteristics}
        \end{subfigure}
        
        \caption{Diverse driving behaviors at an intersection.}
        \vspace{-0.5cm}
        \label{fig:intersection_behaviors}
\end{figure}

\begin{figure*}[t]
        \centering
        % Figure for highway driving behaviors
        \begin{minipage}{0.49\textwidth}
            \centering
            \includegraphics[width=\linewidth]{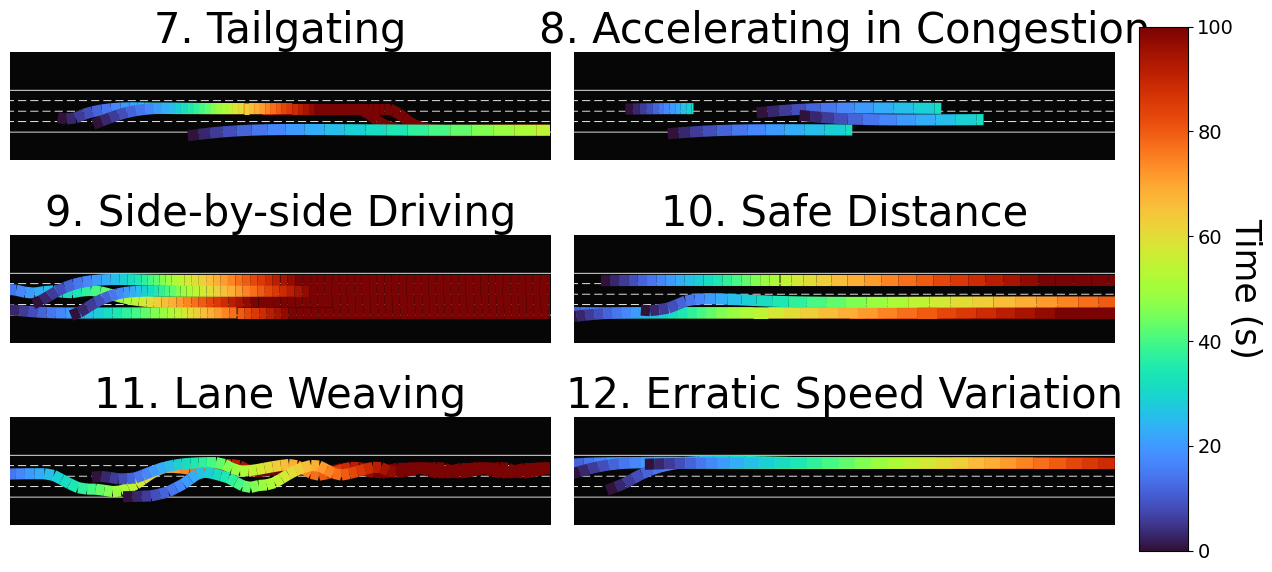}
        %     \caption{Trajectories of highway driving behaviors.}
            \label{fig:highway_traj_visualization}
    
            \vspace{-0.2cm} % Adjust space as needed
            
            \resizebox{\linewidth}{!}{
            \begin{tabular}{lccc}
                \hline
                Behaviors & Emergence Rate (\%) & Collision Rate (\%) & Avg. Speed (m/s) \\ \hline
                \rowcolor[HTML]{EFEFEF} 
                Tailgating & 40.00 & 56.67 & 19.62 \\
                Accelerating in Congestion & 100.00 & 73.33 & 30.18 \\
                \rowcolor[HTML]{EFEFEF} 
                Side-by-Side Driving & 63.33 & 33.33 & 14.28 \\
                Following at a Safe Distance & 93.33 & 0.00 & 31.79 \\
                \rowcolor[HTML]{EFEFEF} 
                Lane Weaving & 73.33 & 36.67 & 15.52 \\
                Erratic Speed & 53.00 & 63.33 & 29.96 \\ \hline
            \end{tabular}}
        %     \caption{Human-annotated emergence rate and driving characteristics.}
            \label{tab:highway_characteristics}
        \end{minipage}
        \hfill
        % Figure for merging behaviors
        \begin{minipage}{0.49\textwidth}
            \centering
            \includegraphics[width=\linewidth]{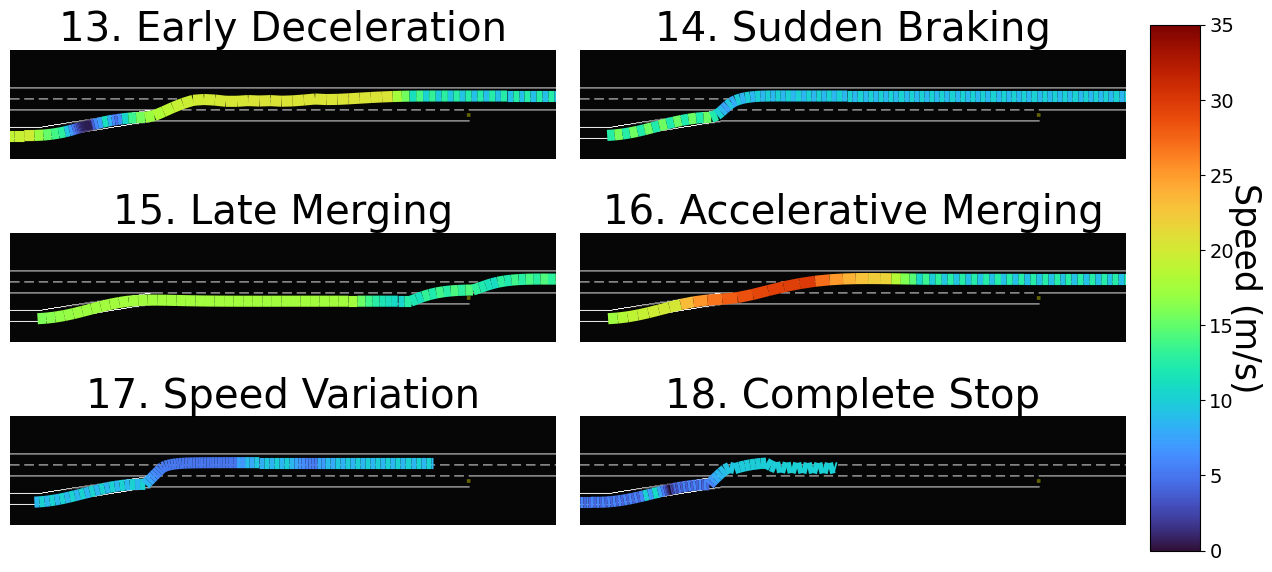}
        %     \caption{Speed profile trajectory of merging behaviors.}
            \label{fig:merge_traj_visualization}
    
            \vspace{-0.2cm} % Adjust space as needed
    
            \resizebox{\linewidth}{!}{
            \begin{tabular}{lccc}
                \hline
                Behaviors & Emergence Rate (\%) & Collision Rate (\%) & Avg. Speed (m/s) \\ \hline
                \rowcolor[HTML]{EFEFEF} 
                Early Deceleration on Ramp & 80.00 & 30.00 & 12.46 \\
                Sudden Braking After Merging & 80.00 & 13.33 & 9.79 \\
                \rowcolor[HTML]{EFEFEF} 
                Late Merging at Ramp End & 86.67 & 0.00 & 13.17 \\
                Accelerative Merging & 60.00 & 33.33 & 12.43 \\
                \rowcolor[HTML]{EFEFEF} 
                Merging with Speed Variation & 96.67 & 6.67 & 8.72 \\
                Merging from Complete Stop & 60.00 & 26.67 & 9.05 \\ \hline
            \end{tabular}}
        %     \caption{Human-annotated emergence rate and driving characteristics.}
            \label{tab:merge_characteristics}
        \end{minipage}
        \caption{Diverse highway driving and merging behaviors.}
        \vspace{-0.5cm}
        \label{fig:driving_behaviors}
\end{figure*}

\section{EXPERIMENTS}
In this section, we demonstrate our method's ability to preserve the behavioral context across natural language, code, and driving policy by showcasing strong alignment between these domains. Following this, we propose a suite of metrics to quantify the behavioral diversity in code and driving policy. Through evaluation, we show that our method can generate more diverse trajectories than other baselines.

\newcommand{\langagreement}{\mathbf{A}^{L \leftrightarrow C}}
\newcommand{\codeagreement}{\mathbf{A}^{C \leftrightarrow P}}
\newcommand{\env}{\mathcal{E}}

\subsection{Implementation Details}
\textbf{Simulator.} We conducted training using the Highway Environment simulator \cite{highway-env}. We simulate for 100 timesteps and update the policy at a frequency of 5Hz. The action space for each vehicle is discrete, comprising 5 possible actions for lateral and longitudinal control with a speed range of $[0,40]$. Our implementation uses the OpenAI Gym framework \cite{brockman2016openai}.

\textbf{Training Details.} Within the MARL framework, we shared rewards for vehicles nearby and penalized for collisions. All policies were trained using the same hyperparameters, network architecture, and environment setup. Specifically, our actor and critic networks each comprise two hidden linear layers, each with 256 neurons followed by a ReLU activation function. We used the RMSprop optimizer and applied a fixed learning rate of $5e^{-5}$ for both networks.

\textbf{Iterating Details.} We train a driving policy $\policypi$, for $10,000$ episodes and do a soft evaluation for every $2,500$ episodes. This uses our iterative process to assess the alignment of $\policypi$ to the behavior $\lang$ without updating the auxiliary function. We terminate the training if the iterator indicates the behavior has been learned. After training for $10,000$ episodes, we run the iterative process again and update the auxiliary function according to the iterator.

\textbf{Evaluation Details.} We evaluate the driving policies on 30 rollouts (varied seeding) using the policy with the highest expected cumulative reward during training.

\subsection{Policy Alignment}
% \begin{figure}[htbp]
%         \centering
%         \includegraphics[width=\linewidth]{Figures/code nl similarity matrix.png}
%         \caption{A diagonal line in the code and language agreement matrix indicates that there is a high similarity between the language description and code, and thus we show that the behavioral context is preserved across these domains.}
%         \label{fig:code_nl_agreement}
% \end{figure}
% \begin{figure}[htbp]
%         % Remove the \centering command
%         % \centering
%         \includegraphics[width=\linewidth]{Figures/code policy agreement matrix.png}
%         \caption{A diagonal line in the code and driving policy agreement matrix indicates that the driving policy that was trained by the reward function was most optimal compared to the other evaluated policies, and therefore we show that the behavioral context is preserved across these domains.}
%         \label{fig:code_policy_agreement}
% \end{figure}

\begin{figure}[t]
        \centering
        \begin{subfigure}[b]{\linewidth}
                \centering
                \includegraphics[width=\linewidth]{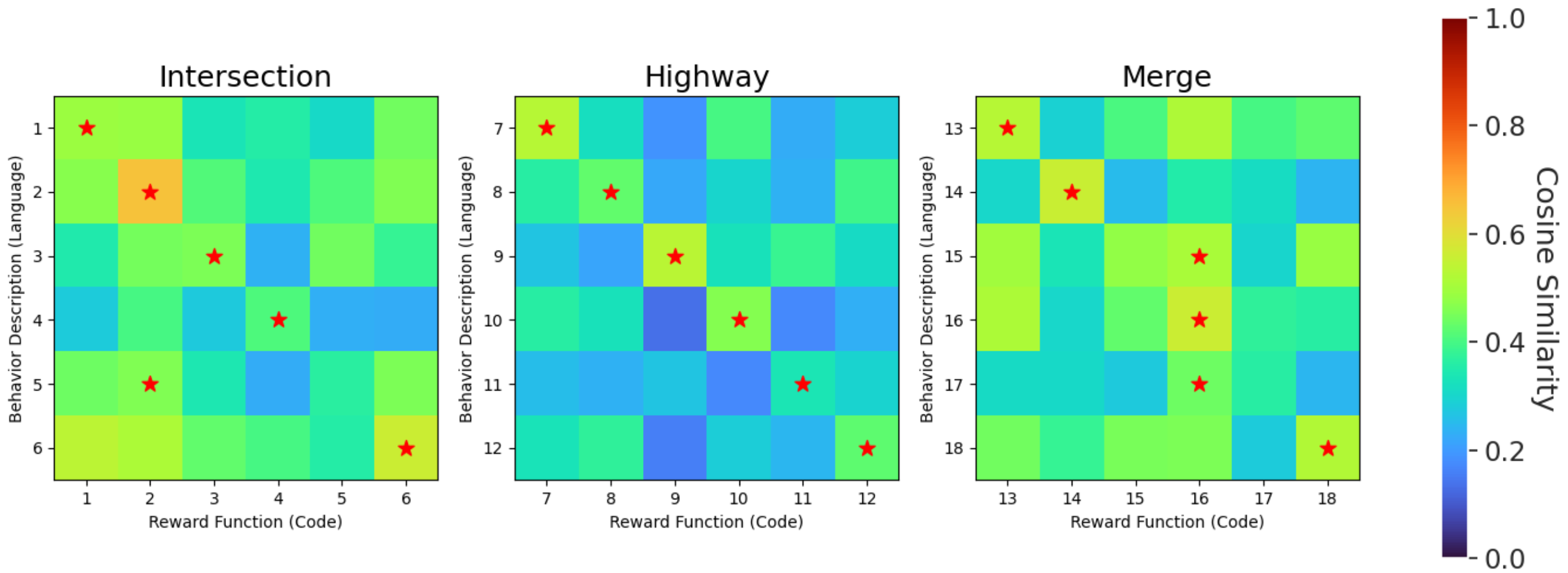}
                \caption{A diagonal line in the code and language agreement matrix indicates that there is a high similarity between the language description and code, and thus we show that the behavioral context is preserved across these domains.}
                \label{fig:code_nl_agreement}
        \end{subfigure}
        \vfill % Adds horizontal space between the subfigures
        \begin{subfigure}[b]{\linewidth}
                \includegraphics[width=\linewidth]{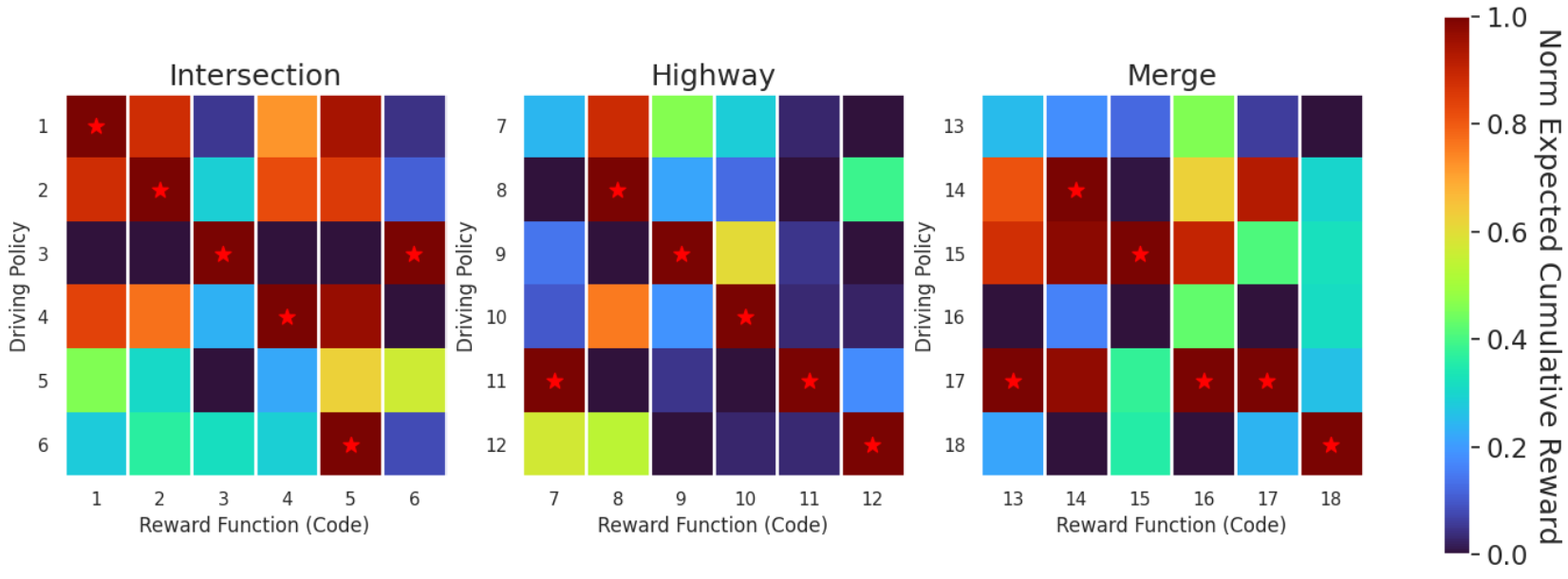}
                \caption{A diagonal line in the code and driving policy agreement matrix indicates that the policy trained by the reward function was most optimal compared to the other evaluated policies, and therefore we show that the behavioral context is preserved across these domains.}
                \label{fig:code_policy_agreement}
        \end{subfigure}
        \caption{Agreement matrix to show behavioral alignment.}
        \vspace{-0.5cm}
        \label{fig:behavioral_alignment}
\end{figure}

In this section, we demonstrate behavioral alignment between the language, code, and driving policy domains using agreement matrices, and validate text-to-driving policy synthesis via manual human inspection.

\textbf{Language and Code Agreement.} To quantify the agreement between the language and code domain, we compute the pairwise cosine similarity between the sentence embedding of $\lang$ and the code embedding of $\auxiliary$ using CodeBERT \cite{feng2020codebert}. For a set of $n$ language descriptions and auxiliary functions, our agreement matrix is a $n \times n$ matrix. A diagonal line in the agreement matrix indicates that the correct pairing of $\lang$ and $\auxiliary$ received the highest similarity among other possible $\lang$ and $\auxiliary$ pairs. \autoref{fig:code_nl_agreement} presents the visualization of the agreement matrix, where a diagonal line is mostly present. Notably, the figure for the ``highway'' environment distinctly shows a pronounced bright diagonal, surrounded by darker regions, indicating a strong alignment. This may be from the greater diversity in language descriptions of the highway behaviors compared to those of other environments.

\textbf{Code and Driving Policy Agreement.} In addition, we extend our analysis to also show strong agreement between code and driving policy domain. To this end, we quantify this agreement by evaluating the trained driving policy $\policypi$ and computing the expected cumulative reward according to the auxiliary reward $\auxiliary$. Specifically, given a set of $n$ auxiliary functions and $n$ driving policies, we define the element $\codeagreement_{ij}$ of the agreement matrix $\codeagreement \in \mathbb{R}^{n \times n}$ as:
\begin{equation}
        \begin{aligned}
                \codeagreement_{ij} = \mathbb{E}_{\tau \sim \policypi_j} \left [\sum_{t=1}^{T} \mathcal{R}_{i}(\lowstate_t, \lowaction_t) \right ]
        \end{aligned}
\end{equation}
Then, we compare these values relative to the performance of alternate driving policies on the same auxiliary reward. Formally, we consider a reward function $\mathcal{R}_{i}$ to be in agreement with a policy $\policypi_j$ if:
\begin{equation}
        \begin{aligned}
                \forall k \neq j, \codeagreement_{ij} > \codeagreement_{ik}
        \end{aligned}
\end{equation}

\begin{figure}[t]
        \centering
        \begin{subfigure}[b]{0.48\linewidth}
            \includegraphics[width=\linewidth]{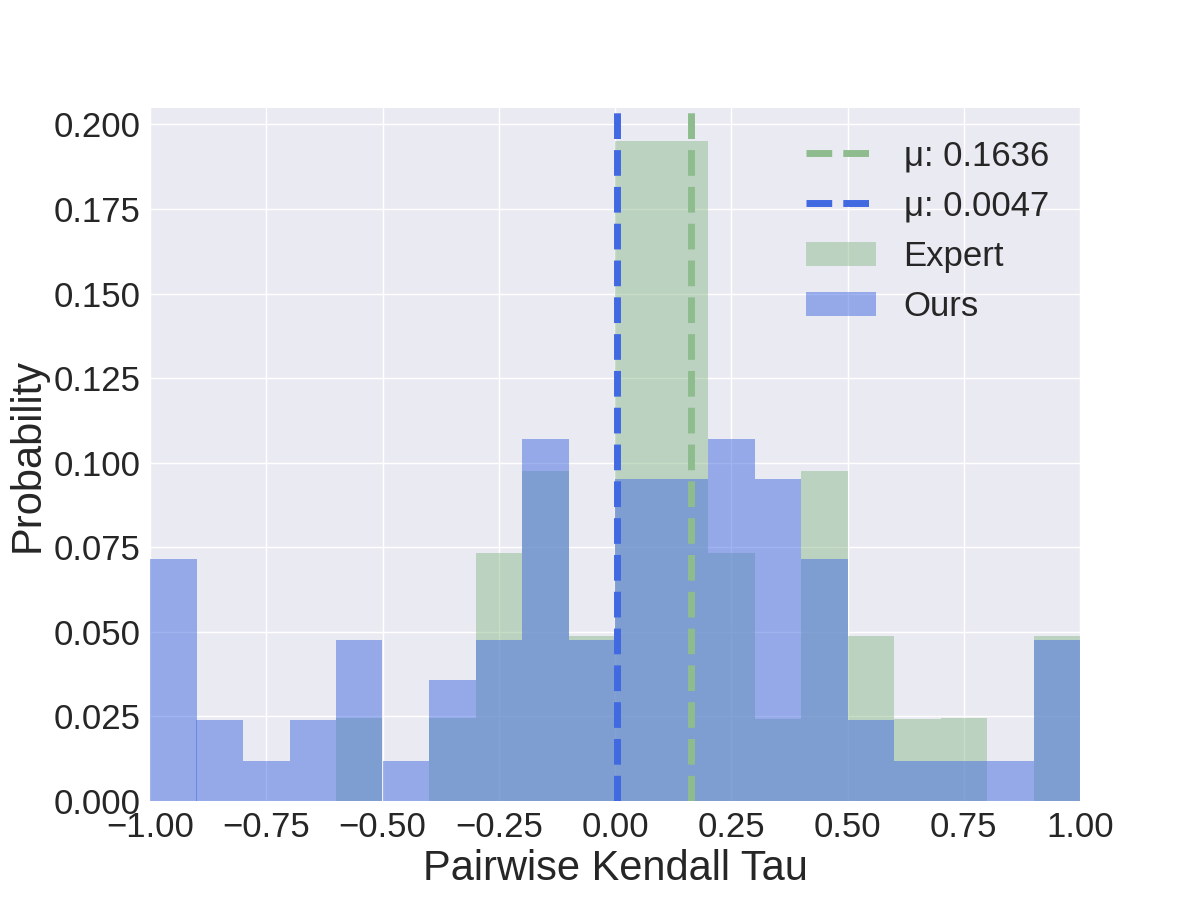}
            \caption{Code diversity with $K=5$}
            \label{fig:code_diversity}
        \end{subfigure}
        \hfill % Adds horizontal space between the subfigures
        \begin{subfigure}[b]{0.48\linewidth}
            \includegraphics[width=\linewidth]{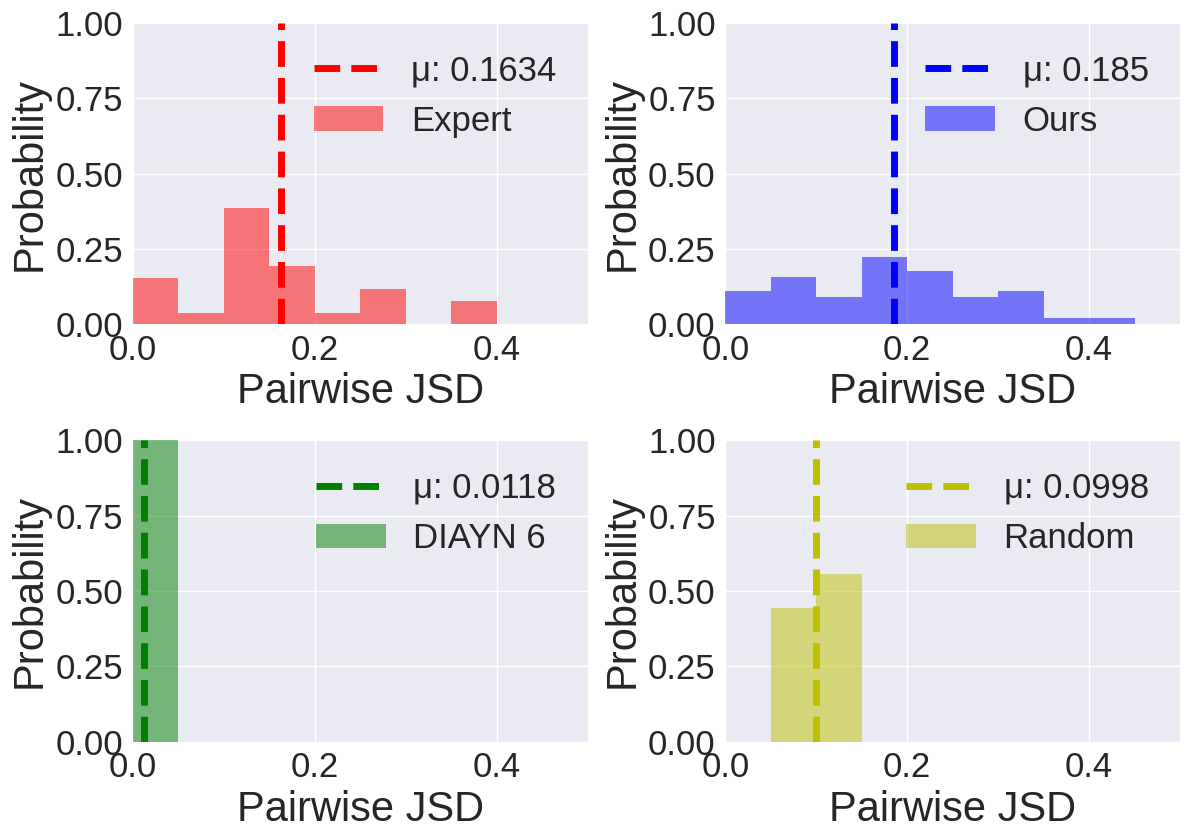 }
            \caption{Policy diversity}
            \label{fig:policy_diversity}
        \end{subfigure}
        \caption{Pairwise diversity comparison.}
        \vspace{-0.5cm}
        \label{fig:baseline_comparisons}
\end{figure}

The intuition is to examine how each driving policy is evaluated by the auxiliary function. The presence of a diagonal line in $\codeagreement$, seen in \autoref{fig:code_policy_agreement} suggests strong alignment between the code and driving policy domain. As observed again, the ``highway'' environment showcases a prominent dark red diagonal that is surrounded by darker blue areas. This implies that the policies are highly specialized and are, therefore, behaviorally diverse. Our results indicate that natural language descriptions may serve as a proxy for estimating the diversity of driving behaviors.

\textbf{Language and Policy Agreement.} In our next analysis, we verify that the driving policy $\policypi$ corresponds to the behaviors described by $\lang$. To evaluate the degree of alignment between the behavior description and the driving policy, we enlist human annotators to measure the emergence rate by manual inspection. The emergence rate measures how frequently the described behavior appears by at least 1 agent in 30 different rollouts. The relatively high emergence rates, as collectively presented in \Cref{tab:intersection_characteristics,tab:highway_characteristics,tab:merge_characteristics}, strongly suggest a consistent adherence to the described behavior rather than stochastic occurrences. The visualizations in \Cref{fig:intersection_traj_visualization,fig:highway_traj_visualization,fig:merge_traj_visualization} showcase this text to driving policy alignment.

\subsection{Diversity Baselines}

\newcommand{\rankreward}{\overline{\mathcal{R}}}
\newcommand{\codediversity}{$\mathcal{D}^{C}$ }
\newcommand{\policydiversity}{$\mathcal{D}^{P}$ }
\newcommand{\Tau}{\mathrm{T}}

\textbf{Code Diversity.} To evaluate the diversity of the reward functions, we utilize the Kendall rank correlation coefficient, $\Tau$, a common statistical metric for assessing the concordance in trends of time series \cite{https://doi.org/10.1002/env.3170030403}, to compare the rank-ordering of rewards. We define the ranked rewards associated with the reward function $\reward$ as $\rankreward$. The ranking process involves discretizing the reward signals into $K$ distinct ranks with $K$ chosen as an odd positive integer to maintain symmetry. Here, a rank of 1 is assigned to the highest positive reward, a middle rank to zero reward, and a rank of $K$ to the lowest negative reward. Ranks 1 to $\frac{K+1}{2}$ categorize positive rewards, while ranks $\frac{K+1}{2}$ to $K$ categorize negative rewards, both distributed into equal parts. We then mask away the central rank when both reward functions exhibit neutral rewards, focusing our analysis on the agreement and disagreement on active reward states. Our diversity metric does not need to assume continuity and is also scale invariant as it measures similarity according to reward prioritization rather than the absolute reward values. We report the median $\Tau$ results at different $K=\{5,7,9\}$ values in \autoref{tab:kendall_tau_baseline}. We will denote this diversity measure as \codediversity.

\begin{table}[t]
        \centering
        % \resizebox{\linewidth}{!}
        {\begin{tabularx}{\linewidth}{X>{\centering\arraybackslash}X>{\centering\arraybackslash}X>{\centering\arraybackslash}X}
                \hline
                \multicolumn{1}{c}{}                                   &                     & \multicolumn{2}{c}{Pairwise Kendall Tau $\downarrow$}                       \\
                \multicolumn{1}{c}{\multirow{-2}{*}{Environment}}              & \multirow{-2}{*}{\begin{tabular}[c]{@{}c@{}}Rank levels\\ ($K$)\end{tabular}} & \multicolumn{1}{c}{Human Expert} & \multicolumn{1}{c}{Ours} \\ \hline
                \rowcolor[HTML]{EFEFEF} 
                \cellcolor[HTML]{EFEFEF}                               & $K=5$                 & -0.0398                    & \textbf{-0.1352}         \\
                \rowcolor[HTML]{EFEFEF} 
                \cellcolor[HTML]{EFEFEF}                               & $K=7$                 & -0.0645                    & \textbf{-0.1189}         \\
                \rowcolor[HTML]{EFEFEF} 
                \multirow{-3}{*}{\cellcolor[HTML]{EFEFEF}Intersection} & $K=9$                 & -0.0538                    & \textbf{-0.1188}         \\
                                                                & $K=5$                 & \textbf{0.2463}            & 0.3675                   \\
                                                                & $K=7$                 & \textbf{0.1985}            & 0.2252                   \\
                \multirow{-3}{*}{Merge}                                & $K=9$                 & \textbf{0.1970}            & 0.3938                   \\
                \rowcolor[HTML]{EFEFEF} 
                \cellcolor[HTML]{EFEFEF}                               & $K=5$                 & 0.0013                     & \textbf{-0.2525}         \\
                \rowcolor[HTML]{EFEFEF} 
                \cellcolor[HTML]{EFEFEF}                               & $K=7$                 & 0.0306                     & \textbf{-0.2358}         \\
                \rowcolor[HTML]{EFEFEF} 
                \multirow{-3}{*}{\cellcolor[HTML]{EFEFEF}Highway}      & $K=9$                 & 0.0351                     & \textbf{-0.2206}         \\ \hline
        \end{tabularx}}
        \caption{Code diversity using Kendall Tau correlation}
        \vspace{-0.2cm}
        \label{tab:kendall_tau_baseline}
\end{table}

\begin{table}[t]
        \centering
        \resizebox{\linewidth}{!}
        {\begin{tabular}{lcccccc}
                \hline
\multicolumn{1}{c}{}                             & \multicolumn{3}{c}{Jensen-Shannon Divergence (IQR) $\uparrow$}                                                        \\ \cline{2-4} 
\multicolumn{1}{c}{\multirow{-2}{*}{Methods}}    & \cellcolor[HTML]{FFFFFF}Intersection & \cellcolor[HTML]{FFFFFF}Merge & \cellcolor[HTML]{FFFFFF}Highway \\ \hline
\rowcolor[HTML]{EFEFEF} 
\cellcolor[HTML]{EFEFEF}Random Policy (6 skills) & 0.1197 (0.0019)                      & 0.2297 (0.0040)               & 0.2515 (0.0022)                 \\
\rowcolor[HTML]{EFEFEF} 
Random Policy (30 skills)                        & 0.1385 (0.0014)                      & 0.2250 (0.0084)               & 0.3033 (0.0007)                 \\
\rowcolor[HTML]{FFFFFF} 
Human Expert (5 skills)                          & 0.1686 (0.0313)                      & 0.2595 (0.0239)               & \textbf{0.3686 (0.0442)}        \\
\rowcolor[HTML]{EFEFEF} 
DIAYN (6 skills)                                 & 0.0107 (0.0062)                      & 0.0152 (0.0038)               & 0.0254 (0.0021)                 \\
\rowcolor[HTML]{EFEFEF} 
DIAYN (18 skills)                                & 0.0163 (0.0039)                      & 0.0211 (0.0058)               & 0.0319 (0.0014)                 \\
\rowcolor[HTML]{EFEFEF} 
DIAYN (36 skills)                                & 0.0181 (0.0079)                      & 0.0083 (0.0027)               & 0.0195 (0.0067)                 \\
\rowcolor[HTML]{FFFFFF} 
Ours (6 skills)                                  & \textbf{0.1845 (0.1085)}             & \textbf{0.3397 (0.0523)}      & 0.3039 (0.0729)                 \\ \hline
                \end{tabular}}
        \caption{Trajectory diversity using JSD}
        \vspace{-0.5cm}
        \label{tab:trajectory_diversity}
\end{table}

A high \codediversity value, close to 1, indicates similar reward rankings and low diversity. Conversely, a value near 0 suggests moderate diversity due to inconsistent reward correlations. Negative \codediversity values, particularly those approaching -1, signify high diversity, as the reward function ranks rewards inversely. In the merge map, a higher similarity is expected due to a common reward that promotes merging. In contrast, other environments displayed lower or negative \codediversity values, attributable to fewer simulation constraints.

We benchmarked our results against the default reward functions from the Highway Env simulator that was developed and refined by the community, which we regarded as expert-crafted rewards \cite{highway-env}. We conducted the same experiments on 5 different expert-craft reward functions per environment and reported the median \codediversity in \autoref{tab:kendall_tau_baseline}. For both the ``intersection'' and ``highway'' environment, \MethodNames consistently yielded lower \codediversity values than those of the expert-crafted reward functions, indicating greater diversity in our reward structures. This discrepancy was most notable in the ``highway'' environment, where \MethodNames exhibited significantly more negative \codediversity values, contrasting that with the positive \codediversity values from the expert reward functions. On the ``merge'' environment, our results indicated less diversity compared to the expert-crafted reward functions, possibly because the expert reward functions did not promote merging behavior as a common reward objective, whereas ours did. Lastly, the histogram in \autoref{fig:code_diversity} shows that \MethodNames resulted in a greater spread of code diversity outcomes, particularly by achieving a higher frequency of negative Kendall Tau correlation values. Moreover, it recorded a lower mean pairwise Kendall Tau score than that associated with expert rewards. This highlights our method's ability to generate varied reward structures to promote diverse driving behaviors.

\textbf{Driving Policy Diversity.} In this section, we show that \MethodNames can also generate behaviorally diverse trajectories. We use an existing metric introduced in \cite{pmlr-v139-lupu21a} to measure the trajectory diversity via the Jensen-Shannon Divergence (JSD). We report the median JSD across all agents on 30 different seedings for each map in \autoref{tab:trajectory_diversity}. 

To contextualize these findings, we benchmarked against three different baselines: random behaviors, unsupervised skill acquisition algorithms, and driving policies trained on expert-crafted reward functions \cite{highway-env}. Random behaviors were generated by defining $\policypi$ as a uniform distribution, where it equally picks an action from the available actions. Then, 30 random behaviors are generated for each map through varied seeding. Next, we compared \MethodNames to Diversity is All You Need (DIAYN) \cite{eysenbach2018diversity}, an established unsupervised skill acquisition algorithm. We adapted the DIAYN method into a multi-agent setting and trained for 3 different skill counts per map (6, 18, and 36). Our third baseline is against driving policies that were trained on expert-crafted reward functions. We report the median \policydiversity in \autoref{tab:trajectory_diversity}. Our results, as summarized in the table, indicate that \MethodNames surpass random policies and DIAYN-generated policies across all tested scenarios. Notably, \MethodNames exhibits the highest \policydiversity in ``merge'' scenarios, suggesting a greater behavioral diversity compared to other methods. Even in ``intersection'' and ``highway'' scenarios, our approach demonstrates competitive diversity, only marginally trailing the human expert in ``highway'' scenarios. The pairwise JSD values distribution shown in \autoref{fig:policy_diversity} suggests that driving policies derived from explicitly defined reward functions tend to yield a more dispersed and wider spread of policy diversity compared to those from intrinsic reward policies and random behaviors. 

\section{ABLATION}

\begin{figure}[htbp]
        % Remove the \centering command
        % \centering
        \includegraphics[width=\linewidth]{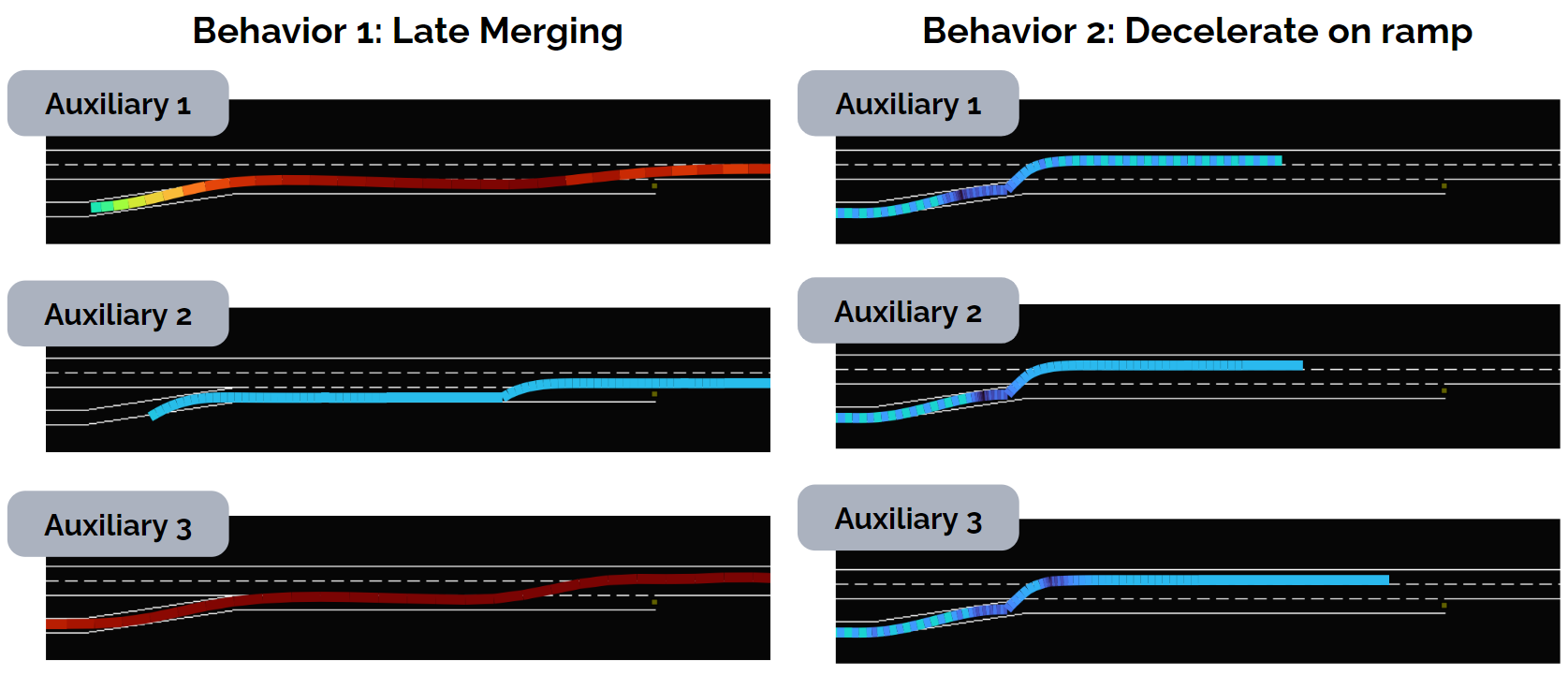}
        \caption{Robustness to auxiliary function generation.}
        \label{fig:auxiliary}
\end{figure}

\textbf{Robustness to Auxiliary Functions.} We performed an ablation study to assess the impact of different auxiliary functions. We trained driving policies using the same primary function but varied the auxiliary function. The results, shown in \autoref{fig:auxiliary}, demonstrate that our method is robust to different generations of the auxiliary function, as they all exhibit similar behavior trajectories. We believe this is because the primary function provides a strong signal for the driving policy to learn the desired behavior, while the auxiliary function provides additional guidance for exploration.

\section{DISCUSSION}

% Limitations: dealing with traffic violations
% Inject human interpetability by defining functions -- it would be better if these states could be determined from perception

In this section, we discuss limitations and promising directions for future work. 

A limitation of our work is that we did not train the driving policies using data-driven simulators, potentially restricting their ability to generalize to real-world scenes. Consequently, these driving policies may struggle with real-world complexities, such as following traffic regulations. For instance, the driving policy may not learn to stop at a red light, as this behavior is not explicitly defined. Another limitation is that we inject human interpretability by defining low-level functions (e.g. \texttt{is\_on\_ramp}). Ideally, these abstract states should be determined through the perception layer using bounding boxes or semantic segmentation.

A promising direction for future work involves training the driving policies on data-driven simulators with an RL interface, such as Waymax \cite{gulino2023waymax}. This approach allows for generating abstract states by querying the semantic map layer. By combining our knowledge-driven approach with data-driven simulators, we can synthesize diverse behaviors for surrounding vehicles, thereby replacing traditional driver models, such as the intelligent driver model.

\section{CONCLUSION}

% practical implications in enhancing the realism of driving interactions 
% reflects the potential of our approach to more diverse AV scenario validation while maintaining realism
% Our results demonstrate that our method can generate diverse and realistic driving policies across different maps.

In our work, we introduce Text-to-Drive (T2D) to generate diverse driving behaviors from natural language descriptions. T2D utilizes an LLM to synthesize behaviors in simulation, constructing a state chart for mapping states to high-level abstractions, thereby enhancing task summarization, policy alignment assessment, and auxiliary reward shaping without human supervision. With our knowledge-driven approach, we demonstrate that T2D generates more diverse trajectories compared to other baselines and offers a natural language interface that allows for incorporating human preference.


%%%%%%%%%%%%%%%%%%%%%%%%%%%%%%%%%%%%%%%%%%%%%%%%%%%%%%%%%%%%%%%%%%%%%%%%%%%%%%%%

%%%%%%%%%%%%%%%%%%%%%%%%%%%%%%%%%%%%%%%%%%%%%%%%%%%%%%%%%%%%%%%%%%%%%%%%%%%%%%%%

%%%%%%%%%%%%%%%%%%%%%%%%%%%%%%%%%%%%%%%%%%%%%%%%%%%%%%%%%%%%%%%%%%%%%%%%%%%%%%%%

% \clearpage

\renewcommand{\sectionautorefname}{App.}
\renewcommand{\subsectionautorefname}{App.}
\renewcommand{\subsubsectionautorefname}{App.}
\setcounter{equation}{0}  % reset counter     
\addtocontents{toc}{\protect\setcounter{tocdepth}{0}}

\bibliographystyle{plain}
\bibliography{./root.bib}

% Start of appendix
\appendices
\onecolumn

\section{FULL PROMPTS}

\noindent \textbf{Prompt 1: State chart prompt}
\begin{shaded}
        \begin{Verbatim}[commandchars=\\\{\}]
System: The AI is talkative and provides lots of specific details from its 
context. If the AI does not know the answer to a question, it truthfully says it 
does not know.

You are a scientist and you are trying to design a statechart for the environment 
that will keep track of the vehicle's state according to the behavior. As an 
example, the statechart function signature is:

def _agent_statechart(self, vehicle) -> None: 
        ...
        return None

Some helpful tips for writing the statechart code:
(1) Use the functions provided in the context. Do not create any new functions. 
But if you must, ensure they are fully defined and implemented without leaving 
any placeholders.
(2) Use 'update_state(self, state_name:str, vehicle: Vehicle, store_value: bool)' 
to define and update vehicle states.
(3) YOU MUST only update a state when a vehicle has reached that state
(4) Only update a state when it has been reached.
(5) Keep track of states that are important for the behavior. 
(6) When defining states relevant to a behavior, it's essential to capture not 
only the states directly involved in the primary action but also those indirectly 
related. This includes preparatory states before the action begins, any variations 
during the action, and consequential states after the action is completed. 
Ensure the definition of states is comprehensive, covering the entire spectrum 
from initial conditions to final outcomes, to provide a full understanding of 
the behavior in its entirety.

You are given the following context from the source code to help you answer 
the question at the end.

Context: \{context\}
Question: \{question\}
Helpful Answer: Let's think step by step.
        \end{Verbatim}
\end{shaded}

\noindent \textbf{Prompt 2: Primary function prompt}
\begin{shaded}
        \begin{Verbatim}[commandchars=\\\{\}]
System: The AI is talkative and provides lots of specific details from its context. 
If the AI does not know the answer to a question, it truthfully says it does not 
know.

Your task is to implement a function tasked with verifying a vehicle's visitation 
of certain states that are crucial for a specific behavior, concluding with the 
vehicle being in a final, key state. This function needs to dynamically recognize 
if the vehicle has visited all necessary states associated with the behavior and 
is currently in the correct concluding state. The sequence of these visits, aside 
from ending in the final state, does not need to follow a predetermined order. 
The function awards a reward of 5 only when it confirms that all relevant states 
have been visited and the vehicle is in the final state, indicating the behavior 
aligns with the criteria set. As an example, the fitness function signature is:

def _agent_fitness(self, vehicle) -> float:

To read the current state of the vehicle, use the method: 
"def _read_state(self, state_name: str, vehicle: Vehicle) -> bool". This method 
determines if the vehicle is currently in a given state. 

To check if the vehicle has visited a state in the past, use the method 
"def _read_visited_state(self, state_name: str, vehicle: Vehicle) --> bool". This 
method returns "True" if the state has been visited, regardless of when it occurred. 

To count the total number of visits to a specific state over time, use the method 
"def _count_total_state_visit(self, state_name: str, vehicle: Vehicle) -> int". 
This method sums up all visits, whether consecutive or not. 

To count how often a vehicle has entered and exited a specific state in consecutive 
visits, use "def _count_state_visit_cycle(self, state_name: str, vehicle: Vehicle) 
-> int". A 'cycle' is defined as a series of consecutive visits to a state, with 
each uninterrupted sequence counted as one cycle. 

These are already defined, you only need to write the fitness function 
_agent_fitness(). Use only these four functions. Do not create any new functions.

Answer the question at the end. Make sure the code output should be formatted as 
a python code string: "'''python ... '''".

Question: \{question\}
Helpful Answer: Let's, think step by step.
        \end{Verbatim}
\end{shaded}

\noindent \textbf{Prompt 3: Auxiliary function prompt}
\begin{shaded}
        \begin{Verbatim}[commandchars=\\\{\}]
System: The AI is talkative and provides lots of specific details from its 
context. If the AI does not know the answer to a question, it truthfully 
says it does not know.

You are a reward engineer tasked with the creation of a reward function 
specifically designed to guide an agent toward a behavior as outlined by 
the task requirements. For behaviors that require visiting specific states, 
structure the reward function to issue rewards only after the agent has 
visited certain key states, like granting a reward for reaching a goal state 
only if the agent has previously visited a critical prerequisite state. 
If you are going to use a variable that does not exist make sure to define 
it. As an example, the reward function signature is:

def _agent_reward(self, vehicle) -> float:
        ...
        reward = max(min(reward, 1.0), -1.0)
        return reward, reward_dict

The code output should be formatted as a python code string: "'''python ... '''".

Some helpful tips for writing the reward function code:
(1) YOU MUST: ensure the total reward is within the range of -1 and 1. 
Achieve this by carefully choosing the reward and penalty values so that you 
incentivize the right behaviors. Avoid using large numbers that may result in 
the reward being outside the range.
(2) The output reward should consist of two items: (1) total reward, (2) a 
dictionary of each reward component.
(3) Provide a detailed incentive structure.
(4) Always round the speed to the nearest integer before comparing it to a value.
(5) Work around the fitness function to provide additional guidance for the 
behavior. But also provide more frequent rewards to help the agent get to the 
state faster. Avoid adding dense rewards that will change the behavior of the 
agent.
(6) Add a collision penalty at the end. To get the collision penalty call 
self.add_collision_penalty(vehicle, collision_penalty). The default collision 
penalty is -0.7. You may change the collision penalty if you think it is 
necessary.
(7) You can add new reward components to help guide the agent to the behavior. 
But make sure the reward is well-balanced and does not change the behavior 
of the agent.

You are given the following context from the source code to help you 
answer the question at the end.

Context: \{context\}
Question: \{question\}
Helpful Answer: Let's think step by step.
        \end{Verbatim}
\end{shaded}

\section{EXAMPLES}

\noindent \textbf{Example: State chart function for the behavior ``Late merging"}
\begin{shaded}
        \begin{Verbatim}[commandchars=\\\{\}]
def _agent_statechart(self, vehicle) -> float:
   # Check if the vehicle is at the acceleration area
   if self._is_at_acceleration_area(vehicle):
        # Update the state
        self._update_state('at_acceleration_area', vehicle, True)
        
        # Check if the vehicle is close to the end of the acceleration area
        if self.distance_to_merging_end(vehicle) <= 30:
        # Update the state
        self._update_state('close_to_end_of_acceleration_area', vehicle, True)
        else:
        # If the vehicle is not close to the end, reset the state
        self._update_state('close_to_end_of_acceleration_area', vehicle, False)
   else:
        # If the vehicle is not at the acceleration area, reset the states
        self._update_state('at_acceleration_area', vehicle, False)
        self._update_state('close_to_end_of_acceleration_area', vehicle, False)
        
   # Check if the vehicle has merged to the highway
   if self._check_merged(vehicle):
        # Update the state
        self._update_state('merged_to_highway', vehicle, True)
   else:
        # If the vehicle has not merged, reset the state
        self._update_state('merged_to_highway', vehicle, False)

   return None        
        \end{Verbatim}
\end{shaded}

\noindent \textbf{Example: Primary function for the behavior ``Late merging"}
\begin{shaded}
        \begin{Verbatim}[commandchars=\\\{\}]
def _agent_fitness(self, vehicle) -> float:
    # Check if the vehicle was at the acceleration area at any point in the past
    visited_acceleration_area = self._read_visited_state('at_acceleration_area', 
                                                          vehicle)
        
    # Check if the vehicle was close to the end of the acceleration area at 
    # any point in the past
    visited_close_to_end_of_acceleration_area = self._read_visited_state
    ('close_to_end_of_acceleration_area', vehicle)
        
    # Check if the vehicle is currently merged to the highway
    currently_merged_to_highway = self._read_state('merged_to_highway', vehicle)
        
    # If all conditions are met, return a reward of 5
    if visited_acceleration_area and visited_close_to_end_of_acceleration_area and 
    currently_merged_to_highway:
        return 5.0
        
    # If not all conditions are met, return a reward of 0
    return 0.0
        \end{Verbatim}
\end{shaded}

\noindent \textbf{Example: Auxiliary function for the behavior ``Late merging"}
\begin{shaded}
        \begin{Verbatim}[commandchars=\\\{\}]
def _agent_reward(self, vehicle) -> float:
        # Initialize the reward and the reward component dictionary
        reward = 0.0
        reward_dict = {}
        
        # Check if the vehicle is at the acceleration area
        at_acceleration_area = self._is_at_acceleration_area(vehicle)
        reward_dict['at_acceleration_area'] = 0.2 if at_acceleration_area else 0.0
        
        # Check if the vehicle is close to the end of the acceleration 
        # area (within 30 meters)
        distance_to_end = self.distance_to_merging_end(vehicle)
        close_to_end_of_acceleration_area = distance_to_end <= 30
        reward_dict['close_to_end_of_acceleration_area'] = 0.5 if 
        close_to_end_of_acceleration_area else 0.0
        
        # Check if the vehicle has merged to the highway
        merged_to_highway = self._check_merged(vehicle)
        reward_dict['merged_to_highway'] = 0.3 if merged_to_highway else 0.0
        
        # Calculate the total reward
        reward += sum(reward_dict.values())
        
        # Ensure the total reward is within the range of -1 to 1
        reward = max(min(reward, 1.0), -1.0)
        
        # Return the total reward and the reward component dictionary
        return reward, reward_dict
        \end{Verbatim}
\end{shaded}

\end{document}